\documentclass[runningheads]{llncs}
\usepackage{graphicx}
\usepackage{algorithm}
\usepackage{algpseudocode}
\usepackage{amsmath}
\usepackage{amsfonts}
\usepackage{booktabs}
\usepackage{subfigure}
\usepackage{multirow}
\usepackage{marvosym}

\begin{document}
%
\title{A Local Similarity-Preserving Framework for Nonlinear Dimensionality Reduction with Neural Networks}
\titlerunning{Vec2vec: A Local Similarity-Preserving Framework}
%
\author{Xiang Wang\inst{1} \and
Xiaoyong Li\inst{1} \and
Junxing Zhu\inst{1} \textsuperscript{\Letter} \and
Zichen Xu\inst{2} \and
Kaijun Ren\inst{1,3} \and
Weiming Zhang\inst{1,3} \and
Xinwang Liu\inst{3} \and
Kui Yu\inst{4}
}
\authorrunning{X. Wang et al.}
%
\institute{College of Meteorology and Oceanography\\
	National University of Defense Technology, Changsha, China \\
\email{\{xiangwangcn,sayingxmu,zhujunxing,renkaijun,wmzhang\}@nudt.edu.cn}
\and
College of Computer Science and Technology\\
Nanchang University, Nanchang, China\\
\email{xuz@ncu.edu.cn}
\and
College of Computer Science and Technology\\
National University of Defense Technology, Changsha, China \\
\email{xinwangliu@nudt.edu.cn}
\and
School of Computer Science and Information Engineering \\
Hefei University of Technology, Hefei, China\\
\email{yukui@hfut.edu.cn}}
%
\maketitle              
\begin{abstract}
Real-world data usually have high dimensionality and it is important to mitigate the curse of dimensionality. High-dimensional data are usually in a coherent structure and make the data in relatively small true degrees of freedom. There are global and local dimensionality reduction methods to alleviate the problem. Most of existing methods for local dimensionality reduction obtain an embedding with the eigenvalue or singular value decomposition, where the computational complexities are very high for a large amount of data. Here we propose a novel local nonlinear approach named \emph{Vec2vec} for general purpose dimensionality reduction, which generalizes recent advancements in embedding representation learning of words to dimensionality reduction of matrices. It obtains the nonlinear embedding using a neural network with only one hidden layer to reduce the computational complexity. To train the neural network, we build the neighborhood similarity graph of a matrix and define the context of data points by exploiting the random walk properties. Experiments demenstrate that \emph{Vec2vec} is more efficient than several state-of-the-art local dimensionality reduction methods in a large number of high-dimensional data. Extensive experiments of data classification and clustering on eight real datasets show that \emph{Vec2vec} is better than several classical dimensionality reduction methods in the statistical hypothesis test, and it is competitive with recently developed state-of-the-art UMAP.

\end{abstract}
\section{Introduction}\label{sec:introduction}
Real-world data, such as natural languages, digital photographs, and speech signals, usually have high dimensionality. Moreover, coherent structure in the high-dimensional data leads to strong correlations, which makes the data in relatively small true degrees of freedom. To handle such high-dimensional real-word data effectively, it is important to reduce the dimensionality while preserving the properties of the data for data analysis, communication, visualization, and efficient storage.

Generally, there are two kinds of methods for dimensionality reduction: one is to preserve the global structure and the other is to preserve the local geometry structure~\cite{mcinnes2018umap,ting2018on}. First, for the dimensionality reduction methods of preserving the global structure of a data set, the most popular methods are Principal Components Analysis (PCA), Linear Discriminant Analysis (LDA), Multidimensional Scaling (MDS), Isometric Feature Mapping (Isomap), Autoencoders, and Sammon Mappings~\cite{van2009dimensionality,cunningham2015linear} and so on. We note that these global methods are either in a strong linearity assumption or can not capture local manifold intrinsic geometry structure~\cite{ting2018on}. Second, for the dimensionality reduction methods of preserving the local geometry structure of a data set, there are some typical methods like Locally Linear Embedding (LLE), Laplace Eigenmaps (LE), Local tangent space alignment (LTSA), t-SNE, LargeVis, and UMAP~\cite{mcinnes2018umap,huang2019dimensionality,kang2020structure}. These methods can learn the manifold intrinsic geometry structure of a data set, which is a very useful characteristic in pattern recognition~\cite{tang2016a}. Most of them share a common construction paradigm: they first choose a neighborhood for each point and then take an eigenvalue decomposition or a singular value decomposition to find a nonlinear embedding~\cite{ting2018on}. \textbf{However, the computational complexities of obtaining an embedding with the eigenvalue decomposition or singular value decomposition ($O(n^2)$) are unbearably expensive~\footnote{Complexity analysis of manifold learning. https://scikit-learn.org/stable/modules/manifold.html}, especially when facing a large number of high-dimensional data.} Table~\ref{tab:comlexity} shows the high computational complexities of four typical local dimensionality reduction methods comparing to \emph{Vec2vec}~\cite{van2009dimensionality,cunningham2015linear,mcinnes2018umap}.

\begin{table}[t]
	\centering
	\caption{Computational complexities of \emph{Vec2vec} and other four state-of-the-art local dimensionality reduction (manifold learning) methods~\cite{van2009dimensionality,cunningham2015linear,mcinnes2018umap}. $n$ is the number of data and $k$ is the number of selected neighbors. $D$ is the input dimensionality and $d$ is the output dimensionality. $|E|$ is the number of edges in the adjacency graph.}
	\begin{tabular}{ccc}
		\hline
		\textbf{Method} & \textbf{Computational} & \textbf{Memory}   \\
		\hline
		LLE   & $O(n\log n\cdot D+n^2\cdot d)$ & $O(|E|\cdot d^2)$  \\
		LE   & $O(n\log n\cdot D+n^2\cdot d))$ & $O(|E|\cdot d^2)$ \\
		LTSA & $O(n\log n\cdot D+n^2\cdot d) $& $ O(n^2)$  \\
		t-SNE & $O(n^2\cdot d) $& $ O(n^2)$ \\
		Vec2vec   & $O(n\log n\cdot D+ n\cdot d)$ & $O(n^2)$\\
		\hline
	\end{tabular}%
	\label{tab:comlexity}%
\end{table}

In recent years, with the success of Word2vec, the embedding representation learning of words~\cite{pennington2014glove}, documents, images, networks, knowledge graphs, biosignals, and dynamic graph are developed and successfully applied~\cite{cui2018survey,cai2018comprehensive,goyal2020dyngraph2vec}. This kind of method transfers the raw data like texts and graphs to low-dimensional numerical vectors or matrices for computing. They implement dimensionality reduction of data, but they are specified in the raw data like texts or graphs~\cite{cui2018survey}, since they utilize the context or structure of the data points for computing. In this kind of methods, Word2vec employs neural networks with a single hidden layer to learn the low-dimensional representation of words~\cite{mikolov2013distributed}. The computational and memory complexities of obtaining an embedding in the method are linear to the number of data. \textbf{Comparing to the eigenvalue or singular value decomposition for obtaining an embedding in many manifold learning methods, it can significantly reduce the computational and memory costs.} The skip-gram model in Word2vec has been successfully applied in the embedding of graphs and networks~\cite{perozzi2014deepwalk,grover2016node2vec}. Word2vec employs the contexts (co-occurrences) of words in texts to learn embedding, but there are no explicit contexts/co-occurrences for the data points in a matrix. Therefore, it cannot be applied to general purpose dimensionality reduction of matrices.


To address these problems, we propose a general-purpose dimensionality reduction approach named \emph{Vec2vec}, which preserves the local geometry structure of high-dimensional data. It combines the advantages of local manifold dimensionality reduction methods and Word2vec. It is not specified in the raw data like texts or graphs and can be applied in the dimensionality reduction of any matrices, and it boosts the computational efficiency of obtaining an embedding simultaneously. To achieve these purposes, we generalize the skip-gram model in Word2vec to obtain the embedding and design an elaborate objective to preserve the proximity of data points. We select the neighbors of the data points to establish a neighborhood similarity graph, and define the contexts of data as the sequences of random walks in the neighborhood similarity graph to solve the objective. We conduct extensive experiments of data classification and clustering on eight typical real-world image and text datasets to evaluate the performance of our method. The experimental results demonstrate that \emph{Vec2vec} is better than several classical well-known dimensionality reduction methods in the statistical hypothesis test. Our method is competitive with recently developed state-of-the-art UMAP but more efficient than it in high-dimensional data. Our \emph{Vec2vec} is more efficient than LLE and LE in a dataset with both a large number of data samples and high-dimensionality.

\section{Related Work}~\label{relatedworks}
\textbf{Embedding Representation Learning.} With the success of Word2Vec model in word representation, embedding representation learning has been widely studied in words, documents, networks, knowledge graphs, biosignals, dynamic graph and so on~\cite{cui2018survey,cai2018comprehensive,goyal2020dyngraph2vec}. Mikolov et al.~\cite{mikolov2013distributed} proposed CBOW and skip-gram models, which were widely used in many embedding methods. Pennington et al.~\cite{pennington2014glove} proposed the GloVe model and learned the embedding representation using matrix decomposition. Bojanowski et al.~\cite{bojanowski2017enriching} proposed the FastText model to enrich word embedding representation with subword information. There are also Doc2Vec, Skip-thoughts, PTE, and Paragram-phrase models to learn the embedding of sentences and documents~\cite{kiros2015skip}. 

There are many embedding representation learning models like TADW, TriNDR, TransE, TransConv, RDF2Vec, MrMine and LINE in different kinds of networks or graphs~\cite{cui2018survey,zhang2018network,cai2018comprehensive}. Some methods like Node2vec, Metapath2Vec, and DeepWalk first find the neighbors of a node using random walks, and then employ the skip-gram model to learn the embedding~\cite{perozzi2014deepwalk,grover2016node2vec}. 

However, existing methods are limited to the specified raw data like words, documents, graphs, and so on. They are designed to utilize the context of texts and the structures of graphs for representation learning of words, documents, and graphs~\cite{bojanowski2017enriching,cui2018survey}. Therefore, they can not be be directly applied to general-purpose dimensionality reduction.

\textbf{Dimensionality Reduction.} Generally, there are two kinds of methods in dimensionality reduction~\cite{mcinnes2018umap,ting2018on}, which preserve the global structure and local geometry structure. In recent years, there are also some methods that employ deep learning to reduce the dimensionality of data~\cite{xu2019review,zhang2018local}.

There are a lot of classical dimensionality reduction methods that focus on the global structure of a data set, such as PCA, LDA, MDS, Isomap, and Sammon Mapping~\cite{cunningham2015linear,van2009dimensionality}. The linear unsupervised PCA optimizes an object that maximizes the variance of the data representation, and there are many successful variants of PCA like Kernel PCA (KPCA) and incremental PCA algorithms~\cite{fujiwara2020an}. The nonlinear Isomap optimizes geodesic distances between general pairs of data points in the neighborhood graph. These global methods construct dense matrices that encode global pairwise information~\cite{ting2018on}.

There are also many local manifold learning methods that discover the intrinsic geometry structure of a data set like LLE, LE, LTSA, t-SNE, LargeVis, and UMAP~\cite{huang2019dimensionality,mcinnes2018umap}. These methods choose neighbors for each data point and obtain a nonlinear embedding from an eigenvalue decomposition or a singular value decomposition~\cite{ting2018on}. The nonlinear LLE preserves solely local properties of the data and considers the high-dimensional data points as a linear combination of their nearest neighbors. UMAP is an effective state-of-the-art manifold learning technology for dimension reduction based on Riemannian geometry and algebraic topology~\cite{mcinnes2018umap}. The computational complexity of UMAP is $O(n^{1.14}\cdot D+k\cdot n)$. Local manifold learning methods usually obtain an embedding from an eigenvalue or a singular value decomposition~\cite{ting2018on} and the computational cost is expensive. 

With the success of auto-encoder, there are some deep learning methods for dimensionality reduction like Local Deep-Feature Alignment (LDFA)~\cite{zhang2018local}, extreme learning machine auto-encoder (ELM-AE)~\cite{kasun2016dimension}, and Deep Adaptive Exemplar AutoEncoder~\cite{shao2016spectral}.

\section{Methodology}~\label{ourmethod}

\subsection{Overview}
The problem of dimensionality reduction can be defined as follows. Consider there is a dataset represented in a matrix $M \in \mathbb{R}^{n \times D}$, consisting of $n$ data points (vectors) $x_i (i\in {1, 2, 3, \cdots, n})$ with dimensionality $D$. In practice, the feature dimension $D$ is often very high. Our purpose is to transfer the matrix $M^{n \times D}$ to a low-dimensional matrix $Z^{n \times d} \in \mathbb{R}^{n \times d} (d\ll D)$ with a function $f$ while preserving the most important information in the matrix $M$. Formally, we can use Equation (\ref{Definition}) to represent the problem.
\begin{equation}\label{Definition}
{Z^{n \times d}=f(M^{n \times D})}
\end{equation}




\begin{figure}
	\centering
	\includegraphics[width=8.5cm]{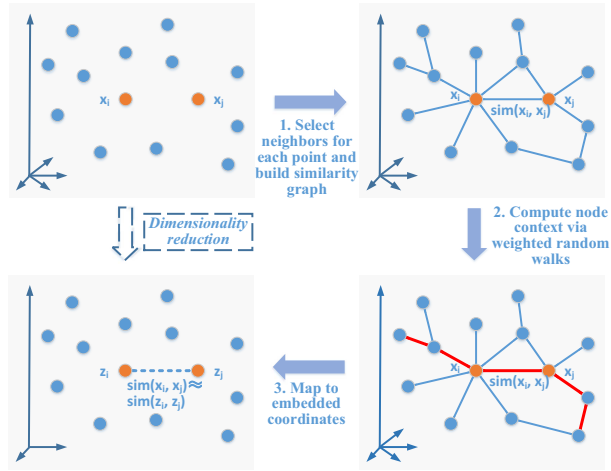}\\
	\caption{The steps of \emph{Vec2vec}: (1) Select nearest neighbors for each data point and construct an adjacency graph. Compute the similarity $sim(x_i, x_j)$ of $x_i$ and its neighbor point $x_j$ as the weight of the edge ($x_i$, $x_j$) in the adjacency graph. (2) Find the contexts of each data point by performing short random walks. The co-occurrences of the data points in the random sequences  reflect their similarity relationships in the neighborhood graph. (3) Compute the low-dimensional embedding vector $z_i$ while preserving the similarities of $x_i$ and its neighbors with neural networks.} \label{Fig_framework}
\end{figure}

\emph{Vec2vec} preserves the pairwise similarities of vectors in a matrix, which are fundamental to machine learning and data mining algorithms. It means that if $sim(x_i, x_j)$ is bigger than $sim(x_a, x_b)$, then $sim(z_i, z_j)$ is going to be bigger than $sim(z_a, z_b)$ in the low-dimensional target space. 

The skip-gram model is originally developed for learning the embedding representation of words in natural languages. We generalize the representation learning model to obtain the embedding. In the skip-gram model, there is a hypothesis that words are likely to be similar if they have similar contexts in sentences. Like words in sentences, our basic hypothesis is that the data points are likely to be similar if they have similar contexts in the feature space. Given the linear nature of texts, it is natural to define the context of words as a sliding window in sentences. However, there are no contexts for data points in a matrix. To solve the problem, we build the neighborhood similarity graph of a matrix and define the context of data points in the matrix as the co-occurrences of data points in the paths of random walks of the graph. 

As shown in Figure~\ref{Fig_framework}, there are mainly three steps in our \emph{Vec2vec} method for dimensionality reduction. The details of the three steps are described in the following subsections.


\subsection{Building Neighborhood Similarity Graph}\label{adjacencyGraph}

To define the the contexts of data points in a matrix and preserve the similarity relationship between the data points, we build an adjacency graph based on their pairwise similarities. We define the similarity graph of a matrix in Definition~\ref{sg}.

\begin{definition}[Similarity Graph ($SG$)]\label{sg}
	\emph{A similarity graph $SG$ of a matrix $M$ is a weighted undirected graph, where the nodes in $SG$ are one-to-one correspondence to the data points in $M$. There are edges between data points and their selected neighbors. The weights of edges are the similarities of the corresponding data points.}
\end{definition}

In Definition~\ref{sg}, given a matrix $M$ with $n$ data points, each data point $x_i$ is represented as a node $v_i$ in $SG$, thus there are $n$ nodes in $SG$. Node $v_i$ and $v_j$ are connected by an edge if $x_i$ ($x_j$) is one of the most similar vectors of $x_j$ ($x_i$). There are two variations to select neighbors for a data point: (a) $\epsilon$-neighborhoods ($\epsilon \in \mathbb{R}$). Node $v_i$ and $v_j$ are connected by an edge if $sim(x_i,x_j)>\epsilon$. This variation is geometrically motivated and the pairwise relationship is naturally symmetric, but it is difficult to choose $\epsilon$. (b) $Topk$ nearest neighbors ($topk \in \mathbb{N}$). Nodes $v_i$ and $v_j$ are connected by an edge if $v_i$ is one of the $topk$ nearest neighbors of $v_j$ or $v_j$ is one of the $topk$ nearest neighbors of $v_i$. This variation is easier to implement and the relation is symmetric, but it is less geometrically intuitive comparing to the $\epsilon$-neighborhoods variation. In our experiments, we choose the $topk$ nearest neighbors variation for building the similar graph.

To compute the edge weight of two nodes, many commonly used similarity/distance functions like Euclidean distance, Minkowski distance, cosine similarity for vectors can be used. In all our experiments, we choose the similarity function ``cosine measure" shown in Equation (\ref{cosine}) to preserve the pairwise similarity relationships of the vectors in $M$.
\begin{equation}\label{cosine}
sim(v_i,v_j)=sim(x_i,x_j)=(x_i \cdot x_j)/(\|x_i\| \cdot \|x_j\|)
\end{equation}

The computational complexity of building the similarity graph is $O(n^2)$. We employ the K-Nearest Neighbor method with a ball tree to build the similarity graph and the computational complexity of this step can be reduced to $O(nlog(n)\cdot D)$. In a distributed system, this step can be further speed up using a parallel method since we only need to compute the pairwise similarities or distances of data vectors in $M$. In the case of $n\gg D$, we can further reduce the computational complexity of building similarity graph to $O(Dlog(D)\cdot n)$, if we build the neighborhood graph with the transpose of $M$ ($M\in \mathbb{R}^{n \times D}$). In this case, if $Z$ ($Z\in \mathbb{R}^{D \times d}$) is the target low-dimensional matrix  of $M^T$, we can get the target low-dimensional matrix $\acute{Z}$ ($\acute{Z}\in \mathbb{R}^{n \times d}$) of $M$ as $\acute{Z}=M\cdot Z$.

\subsection{Node Context in Similarity Graphs}\label{RandomizedProcedure}


It is natural to define the context of words as a sliding window in sentences. However, the similarity graph is not linear and we need to define the notation of the contexts of data points in the similarity graph. We use random walks in the similarity graph to define the contexts of data points. Random walks have been used as a similarity measure for a variety of problems in the content recommendation, community detection, and graph representation learning~\cite{grover2016node2vec,perozzi2014deepwalk}. The detection of local communities motivates us to use random walks to detect clusters of data points, and the random walk model is effortless to parallelize and several random walkers can simultaneously explore different parts of a graph. Therefore, we define the context of a data point in the similarity graph in Definition~\ref{defContext} based on random walks. With the definition, we define the data points around a data point in the random walk sequences as its contexts. With the linear nature of the sequences, we define the context of data points as a sliding window in the sequences.

\begin{definition}[Node Context in Similarity Graphs]\label{defContext}
	\emph{The node context of a data point in similarity graphs is the parts of a random walk that surround the data point.}
\end{definition}

Formally, let $(x_{w1}, x_{w2}, \cdots, x_{wl})$ denote a random walk sequence with length $l$. We use a small sliding window $c$ to define the context of a data point. Then given a data point $x_{wj}$ in the random walk sequence, we can define its node context $NC(x_{wj})$ in Equation (\ref{nodecontext}).
\begin{equation}\label{nodecontext}
NC(x_{wj})=\{x_{wm}| -c\leq m-j \leq c, m\in (1, 2, \cdots, l)\}
\end{equation}

The random walk sequences in similarity graphs can be defined as follows. A random walk is a Markov chain, and the $t$-$th$ data point only depends on the $(t-1)$-$th$ data point in a random walk. The $t$-$th$ data point $x_{wt}$ is generated by the probability distribution defined in Equation (\ref{equRandomwalk}).

\begin{equation}\label{equRandomwalk}
P(x_{wt}=v_a|x_{w(t-1)}=v_b)=\left\{
\begin{array}{lr}
\frac{sim(v_a,v_b)}{\mathbb{Z}},\:if\:(v_a,v_b)\in E, & \\
0,\: otherwise &
\end{array}
\right.
\end{equation}

where $E$ is the edge set of the similarity graph and $sim(v_a,v_b)$ is the edge weight of $v_a$ and $v_b$ defined in Equation (\ref{cosine}). $\mathbb{Z}$ is the normalizing constant and $\mathbb{Z}=\sum_{(v_b,v_i)\in E} sim(v_b,v_i)$. For each data point in the similarity graph, we simulate a fixed number of random walks. For each random walk, we simulate it in a short fixed-length $l$. Our method ensures that every data point in $M$ is sampled to the node contexts of data points.


In this paper, we assume that there are no rare vectors whose similarities with other vectors are all too small to consider. It means that there are no isolated nodes in the similarity graph. This assumption is acceptable if the dataset is not too small or too sparse, such as the image and text datasets in our experiments.

\subsection{The Low-dimensional Embedding Representation}\label{LearningRepresentation}

\begin{figure}
	\centering
	\includegraphics[width=7cm]{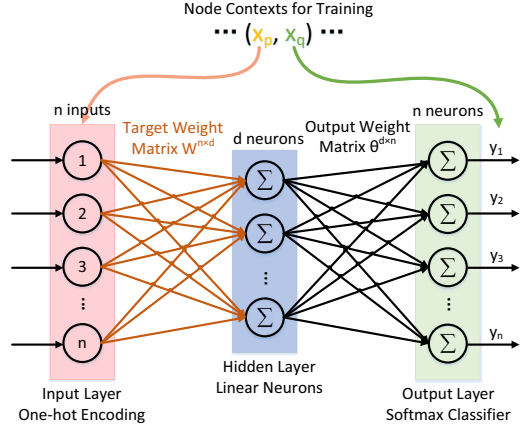}\\
	\caption{The architecture of learning the low-dimensional embedding representation of metrics based on the skip-gram model. It is trained by the data point pairs in the node contexts of the similarity graph.} \label{skip-gram}
\end{figure}

Based on the node contexts of the data points in the similarity graph, we extend the skip-gram model to learn the embedding of the data in the matrix. The skip-gram model aims to learn continuous feature representations of words by optimizing a neighborhood preserving likelihood objective~\cite{mikolov2013distributed}. It is developed to learn the similarity of words from texts by utilizing the co-occurrence of words. The architecture of learning the low-dimensional embedding representation of metrics based on the skip-gram model is shown in Figure~\ref{skip-gram}. It is a neural network with only one hidden layer and the goal of this network is to learn the weight matrix $W$ of the hidden layer, which is actually the target embedding matrix $Z$ ($Z=W$) of the original high-dimensional matrix $M$. 

We use one-hot encoding to represent the data points in the input layer. The data point $x_i$ is represented as a vector $o_i \in \mathbb{R}^n$ in the one-hot encoding, where all elements are zero except the $i$-th element being one. It means that we can get the low-dimensional representation $z_i$ of $x_i$ using equation $z_i= o_i\cdot W$, and the output of the hidden layer of the neural network is $z_i=f(x_i)$. Given $W=[w^1,w^2,\cdots,w^n]\in \mathbb{R}^{n\times d}$, then $z_i$ can be represented as Equation (\ref{equWeight}).
\begin{equation}\label{equWeight}
z_i=f(x_i)= o_i\cdot W=w_i
\end{equation}

The output layer is a softmax regression classifier. The input of this layer is the target embedding vector $z_i$ of $x_i$. The output of this layer is the probability distribution of all data points with the input $x_i$. The neural network is trained by the data pairs in the node contexts of the similarity graph defined in Section~\ref{RandomizedProcedure}. We formulate the dimensionality reduction of matrix $M$ as a maximum likelihood optimization problem. The objective function we seek to optimize is shown in Equation (\ref{ObjFunction}), which maximizes the log-probability of observing the node context $NC(x_i)$ of the data point $x_i$ ($x_i \in M$) conditioned on its feature representation $z_i=f(x_i)$ with the mapping function $f: M^{n \times D}\longrightarrow \mathbb{R}^{n \times d}$ in Equation (\ref{ObjFunction}). $NC(x_i)$ is defined in Equation (\ref{nodecontext}). We introduce the neighborhood similarity graph and node context to compute the $NC(x_i)$, which is different from the original skip-gram model.
\begin{equation}\label{ObjFunction}
\max \limits_{f} \sum_{x_i\in M}logPr(NC(x_i)|f(x_i))
\end{equation}

To optimize the Equation (\ref{ObjFunction}), we assume that the likelihood of observing a neighborhood data point is independent of observing any other neighborhood data points given the representation of the source. Hence, the objective function of Equation (\ref{ObjFunction}) can be changed to Equation (\ref{ObjFunctionAssu1}).
\begin{equation}\label{ObjFunctionAssu1}
\max \limits_{f} \sum_{x_i\in M}log\prod \limits_{x_j\in NC(x_i)}Pr(x_j|f(x_i))
\end{equation}

In Equation~\ref{ObjFunctionAssu1}, the data point pair $(x_i, x_j)$ ($x_j\in NC(x_i)$) is used to train our model. Let $\theta =[\theta _1^T, \theta _2^T, \cdots, \theta _n^T]^T$ be the output weight matrix and $\theta _j^T$ ($\theta _j \in \mathbb{R}^{d}$) be the $j$-th columns of $\theta$, then $\theta _j^T$ is the corresponding weight vector of the output data point $x_j$. As shown in Figure~\ref{skip-gram}, we employ the softmax function to compute $Pr(x_j|f(x_i))$ in Equation~\ref{ObjFunctionAssu1}. Then $Pr(x_j|f(x_i))$ can be calculated as Equation (\ref{ObjFunctionAssu3}).
\begin{equation}\label{ObjFunctionAssu3}
Pr(x_j|f(x_i))=\frac{exp(\theta _j^T\cdot f(x_i))}{\sum_{x_m\in M}exp(\theta _m^T\cdot f(x_i))}
\end{equation}

We find that $Pr(x_j|f(x_i))$ is expensive to compute for a large dataset, since the computational cost is proportional to the number of data. Therefore, we approximate it with negative sampling for fast calculation. Let $P_n(x)$ be the noise distribution to select negative samples and $k$ be the number of negative samples for each data sample, then $Pr(x_j|f(x_i))$ can be calculated using Equation (\ref{ObjNegative}). 
\begin{equation}\label{ObjNegative}
Pr(x_j|f(x_i))=\sigma (\theta _j^T\cdot f(x_i)) \prod \limits_{neg=1}^{k} {\mathbb{E}_{x_{neg}\sim P_n(x)}[\sigma(-\theta _{neg}^T\cdot f(x_i))]}
\end{equation}
where $\sigma (w)=1/(1+exp(-w))$. Empirically, $P_n(x)$ can be the unigram distribution raised to the 3/4rd power ~\cite{mikolov2013distributed}. In our experiments, we use the toolkit \textit{Gensim} to implement the negative sampling and the sample threshold is set to be 0.001. As a result, given $f(x_i)= w_i$ in Equation (\ref{equWeight}), the objective function for learning the embedding representation can be written as Equation (\ref{ObjFunctionAssu2}). 

\begin{equation}\label{ObjFunctionAssu2}
\begin{split}
\max \limits_{f} \sum_{x_i\in M} \sum_{x_j\in NC(x_i)}[log\sigma (\theta _j^T\cdot w_i)+ 
\sum_{neg=1}^{k} {\mathbb{E}_{x_{neg}\sim P_n(x)} log\sigma(-\theta _{neg}^T\cdot w_i)}]
\end{split}
\end{equation}

To alleviate the problem of over fitting, we add the L2 normalization to the objective function. We finally minimize the objective function $J(W,\theta)$ in Equation (\ref{ObjSGD}).
\begin{equation}\label{ObjSGD}
\begin{split}
J(W,\theta)= &-\frac{1}{n}\{\sum_{x_i\in M} \sum_{x_j\in NC(x_i)}[log\sigma (\theta _j^T\cdot w_i)+ \\
&\sum_{neg=1}^{k} {\mathbb{E}_{x_{neg}\sim P_n(x)} log\sigma(-\theta _{neg}^T\cdot w_i)}]\}+ \frac{\lambda}{2} (||W||_2+||\theta||_2)
\end{split}
\end{equation}

We finally solve Equation (\ref{ObjSGD}) with stochastic gradient descent (SGD) to train the neural network.

\begin{figure}[!t]
	\centering
	\subfigure[Running time with the change of the number of data]{%
		\label{figruntime1a}
		\includegraphics[width=5.5cm]{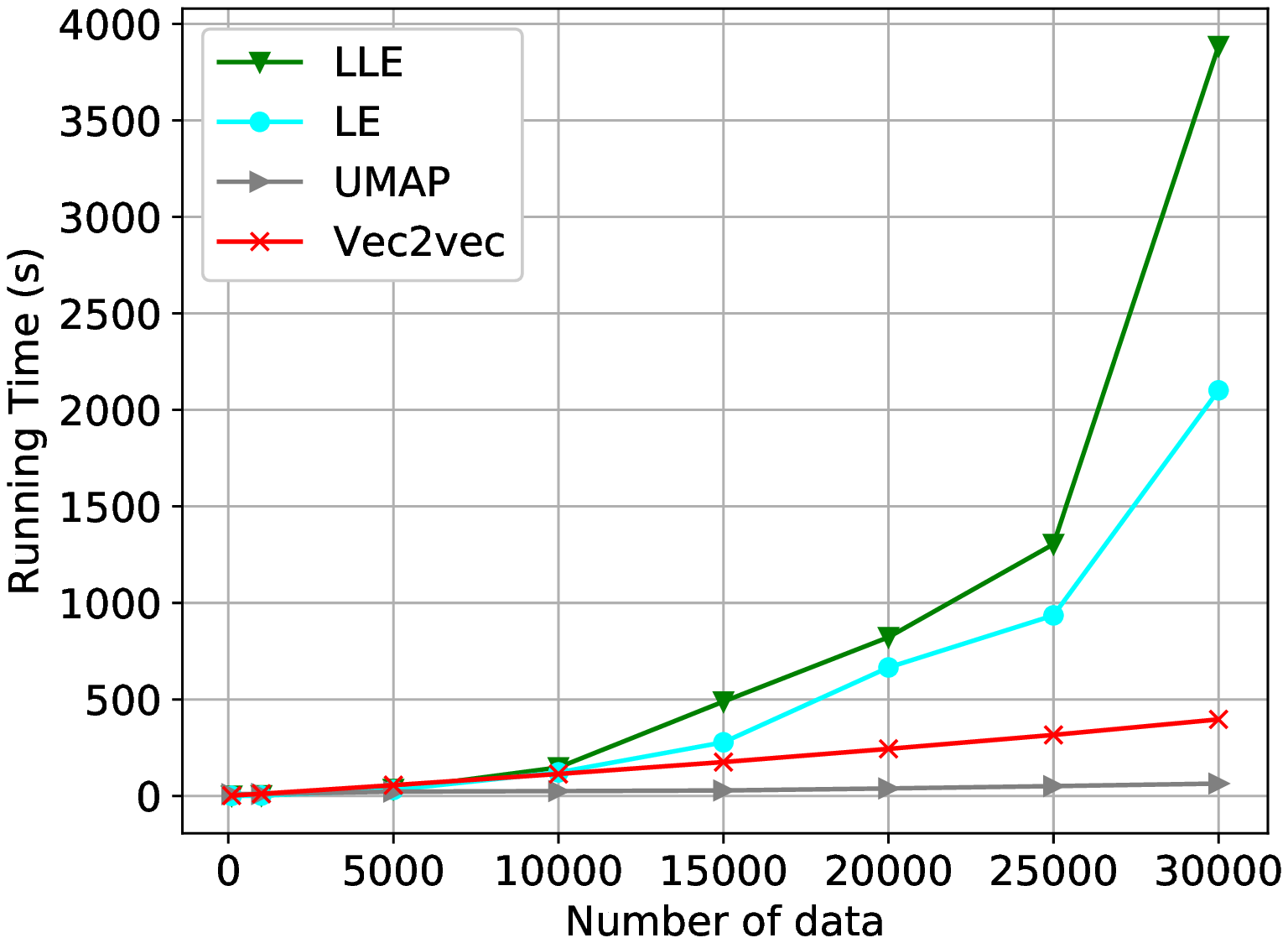}
	}
	\quad
	\subfigure[Runing time with the change of the dimensionality of data]{%
		\label{figruntime1b}
		\includegraphics[width=5.5cm]{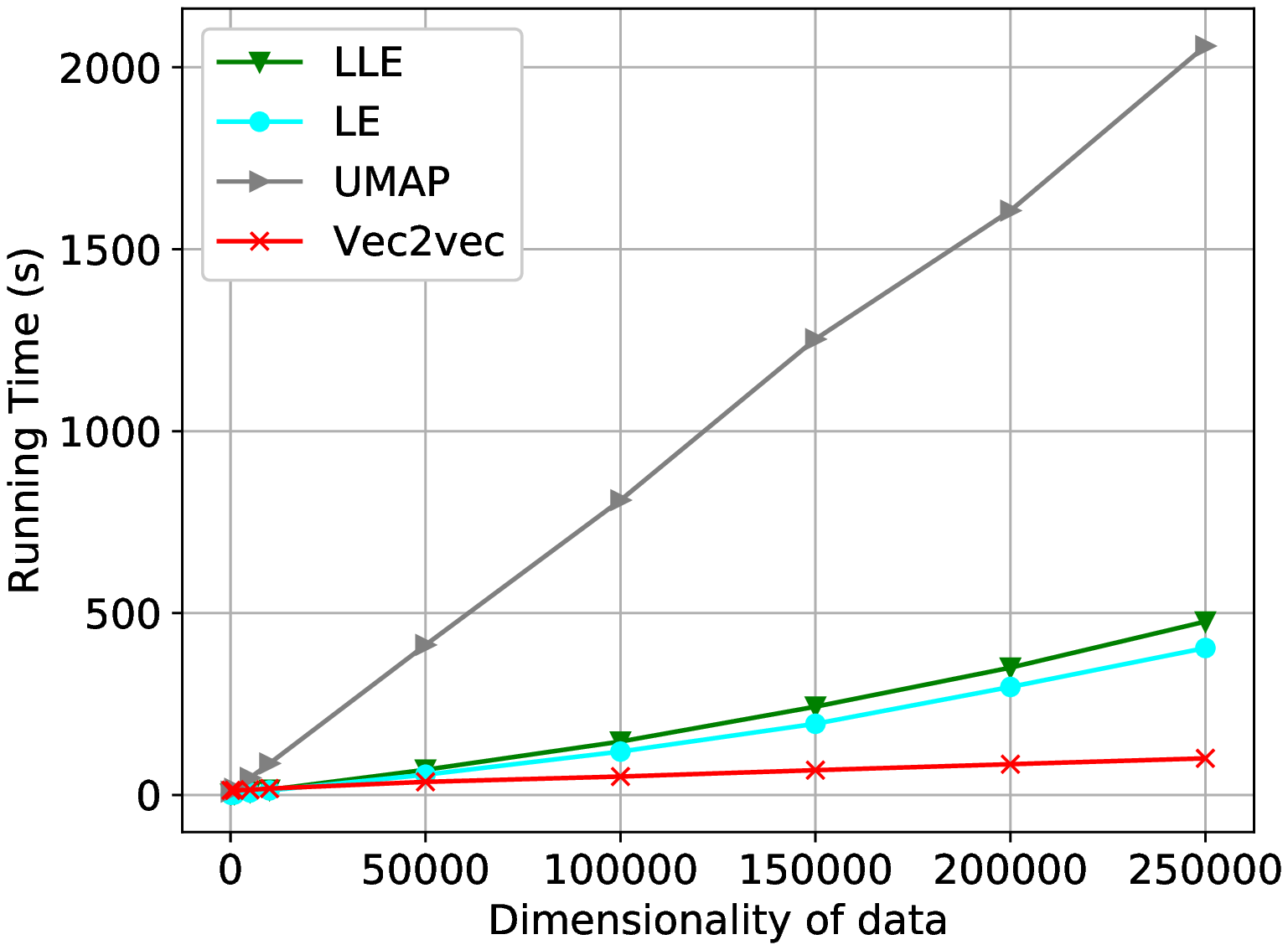}
	}
	\caption{The computational times of the four local dimensionality reduction methods with the change of the number and dimensionality of data. (a) The dimensionality is fixed to $3,072$. (b) The number of data points is fixed to $2,000$.}
	\label{figruntime}
\end{figure}

\section{Experiments}~\label{experiments}

\subsection{Experimental Setup}~\label{expSetup}

For all the classification and clustering tasks, we compare the performance of \emph{Vec2vec} with the six unsupervised methods: (1) PCA, (2) CMDS, (3) Isomap, (4) LLE, (5) LE, and (6) UMAP~\cite{mcinnes2018umap}. PCA, CMDS, and Isomap are typical global methods, while LLE, LE, and UMAP are typical local methods. We use the implementations of the first five methods in the ``scikit-learn" toolkit for experiments, and the implementation of UMAP in Github~\footnote{UMAP in Github. https://github.com/lmcinnes/umap}. To get the best performance of these methods, we set the number of neighbors for each point of Isomap, LLE, LE, and Umap be range from 2 to 30 with step 2 in the experiments. According to~\cite{mcinnes2018umap}, UMAP is significantly more efficient than t-SNE and LargeVis when the output dimensions are larger than 3. Therefore we only compare to UMAP in this paper. We do not compare our method with embedding representation learning methods like Word2vec, Doc2vec, Node2vec, Deepwalk, and LINE~\cite{tang2015line,grover2016node2vec,perozzi2014deepwalk} since they are specific in words, documents, graphs or networks. They cannot be adaptive to the general purpose dimensionality reduction.


\begin{table}[t]
	\newcommand{\tabincell}[2]{\begin{tabular}{@{}#1@{}}#2\end{tabular}}
	\centering
	\caption{The details of the eight text and image datasets used in our experiments}
	\begin{tabular}{c|ccc}
		\toprule
		\tabincell{c}{} & \tabincell{c}{Dataset Name} & \tabincell{c}{Number of data \quad}  &  \tabincell{c}{Dimensionality} \\
		\midrule
		\multicolumn{1}{c|}{\multirow{4}[3]{*}{\tabincell{c}{Image Dataset}}} 
		& MNIST & 5,000 & 784 \\
		& Coil-20 & 1,440 & 1,024 \\
		& CIFAR-10 & 5,000 & 3,072 \\
		& SVHN & 5,000 & 3,072 \\
		\midrule
		\multicolumn{1}{c|}{\multirow{4}[3]{*}{\tabincell{c}{Text Dataset}}}
		& Movie Reviews & 5,000 & 26,197 \\
		& Google Snippets   & 5,000 & 9,561 \\
		& 20 Newsgroups    & 2,000 & 374,855 \\
		& 20 Newsgroups Short   & 2,000 & 13,155 \\
		\bottomrule
	\end{tabular}%
	\label{tab:datasets}%
\end{table}%

In the experiments of classification and clustering, we select four typical real-world image datasets from a variety of domains as shown in Table~\ref{tab:datasets}. For computational reasons, we randomly select 5,000 digits of the SVHN dataset, the CIFAR-10 dataset and the MNIST dataset for our experiments like~\cite{van2009dimensionality}. We represent each images in the datasets as a vector. To test the performance of \emph{Vec2vec} in high-dimensional data, we select four typical text datasets as shown in Table~\ref{tab:datasets}. For the 20 Newsgroups short dataset, we only select the title of the articles in the 20 Newsgroups dataset. In pre-process, we represent each image or text to a vector. We perform some standard text preprocessing steps like stemming, removing stop words, lemmatization, and lowercasing on the datasets. We employ the ``TFIDF" method to compute the weights of the words.

\subsection{Computational Time}\label{secTime}

We compare the computational time of \emph{Vec2vec} with the three local state-of-the-art dimensionality reduction methods with the change of the number and dimensionality of data. As shown in Figure~\ref{figruntime1a}, the computational time of UMAP grows slowest with the growth of the number of data points, and the computational time of \emph{Vec2vec} is the second slowest. As we know, UMAP first constructed a weighted k-neighbor graph and then learned a low dimensional layout of the graph. The first step needs most of the computational time and UMAP optimizes it with an approximate nearest neighbor descent algorithm~\cite{mcinnes2018umap}, while the implementation of our \emph{Vec2vec} did not use a approximate algorithm (the quick approximate algorithm can also be used in our method). So \emph{Vec2vec} is understandable to be a little slower than UMAP. The computational times of LLE and LE are smaller than UMAP when the input number of data is less than 2,000, but the running times of them increase sharply with the growth of the number of data points. The results show the computational efficiency of \emph{Vec2vec} and UMAP in the local dimensionality reduction of large scale of data.

Figure~\ref{figruntime1b} shows the computational time of the four dimensionality reduction methods with the change of the input dimensionality of data. We can find that \emph{Vec2vec} needs the least time when dimensionality reaches nearly 20000. When the dimensionality reaches 100,000, \emph{Vec2vec} needs less computational time than LLE, and the times of the two methods are 146.91 and 140.89. When the dimensionality reaches 150,000, \emph{Vec2vec} needs nearly the same computational time with LLE. The experimental results show that \emph{Vec2vec} is more suitable for dimensionality reduction of high-dimensional data than UMAP, LLE and LE.

\textbf{In summary, UMAP is scalable to a large number of data, but is sensitive to the growth of data dimensionality, while \emph{Vec2vec} is efficient in both a large number of data and high-dimensional data. LLE and LE get better computing performance in a small amount of data and low-dimensional data.}

\subsection{Data Classification}\label{secImageClassification}

%

\begin{figure}[!t]
	\centering
	\subfigure[Low-dimension image data($10^3$)]{%
		\label{Fig_imageClassification}
		\includegraphics[width=5.5cm]{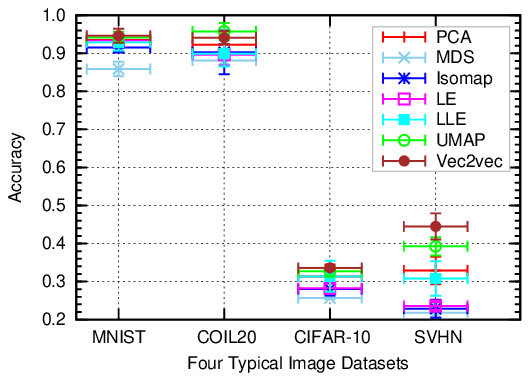}
	}
	\quad
	\subfigure[High-dimension text data($10^4$-$10^5$)]{%
		\includegraphics[width=5.5cm]{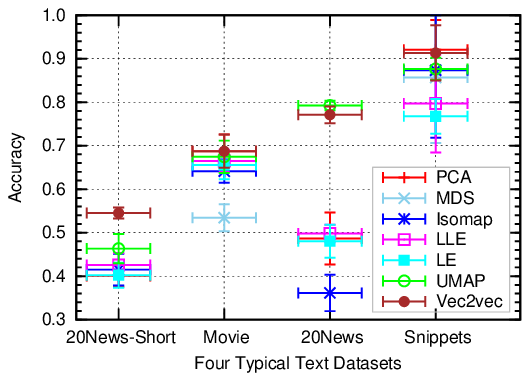}
		\label{Fig_textClassification}
	}
	\caption{The accuracy of the methods on image and text classification.}
	\label{Classification}
\end{figure}

For all the classification experiments, we employ KNN (K-Nearest Neighbor) as our classifier like~\cite{van2009dimensionality} and use the implementation of KNN in ``Scikit-learn". For the choice of parameter $k (k=1,3,5,7,9,11)$ in KNN, we use ``GridSearch" to find the best parameter $k$ in all datasets. We use 4-fold cross-validation to test the performances of different methods and use the accuracy as the performance measure in all the classification experiments. We use the mean accuracy and the 95\% confidence interval of the accuracy estimate (2 times the standard deviation) to be the performance measures.

UMAP is competitive with \emph{Vec2vec} as shown in Figure~\ref{Classification}. We assume that the results are in Gaussian distribution and employ the Student's t-test (alpha=0.1) to test the significant difference between the two methods in the eight datasets. The H-value and p-value are 0 and 0.1543. \textbf{Therefore, UMAP is as good as \emph{Vec2vec} in statistics and the performances of the two methods have no significant difference in the eight datasets.} The skip-gram model used in \emph{Vec2vec} is originally developed to learn the embedding of words while preserving the similarities. The results show that it can be generalized to obtain the embedding while preserving the similarity of data points, which are important for data classification. For the local LLE and LE methods, the H-values are both 1 and the p-values are 0.0195 and 0.0122. \textbf{Therefore, \emph{Vec2vec} is significant better than LLE and LE in the eight datasets in statistics.} For the global PCA, CMDS, and Isomap method, the p-values are 0.0819, 0.0219, and 0.0372. The H-values are all 1. \textbf{Therefore, \emph{Vec2vec} is significantly better than the global PCA, CMDS, and Isomap in the eight datasets in statistics.}

\subsection{Data Clustering}\label{imageClustering}

For all the clustering experiments, we employ spectral clustering with kernel RBF (Radial Basis Function) for clustering, since it is one of the best clustering methods~\cite{von2007tutorial}. We use the implementation of spectral clustering in the ``scikit-learn" library. In our experiments, we set the number of clusters for spectral clustering to be the number of classes in the datasets and set the range of hyperparameter $gamma$ to be from ${10}^{-6}$ to ${10}^{1}$. We use the ``Adjusted Rand Index (ARI)" to be the evaluation metric.


\begin{table}[htbp]
	\newcommand{\tabincell}[2]{\begin{tabular}{@{}#1@{}}#2\end{tabular}}
	\centering
	\caption{The performance of the methods on data clustering. The evaluation metric is ``Adjusted Rand Index (ARI)". The larger the value, the better the method.}
	\begin{tabular}{c|ccccc|cccc}
		\toprule
		\tabincell{c}{Properties\\ to Preserve} & \tabincell{c}{Method}  & \tabincell{c}{MNIST} & \tabincell{c}{COIL20} & \tabincell{c}{CIFAR-10} & \tabincell{c}{SVHN} &  \tabincell{c}{Snippets} &  \tabincell{c}{20News-Short} &  \tabincell{c}{20News} &  \tabincell{c}{Movie} \\
		\midrule
		\multicolumn{1}{c|}{{\tabincell{c}{Global Structure}}\multirow{4}[9]{*}} 
		& PCA   & 0.3682 & 0.6269 & 0.0598 & 0.0185 & 0.0070 & 0.0140 & 0.0421 & 0.0016 \\ 
		& CMDS   & 0.3747 & 0.6472 & 0.0579 & 0.0002 & 0.1007 & 0.0010 & 0.0040 & 0.0006 \\ 
		& ISOMAP & 0.5136 & 0.567 & 0.0533 & 0.0063 & 0.0056 & 0.0178 & 0.0892 & 0.0216 \\
		\midrule
		\multicolumn{1}{c|}{{\tabincell{c}{Local Structure}}\multirow{4}[9]{*}} 
		& LLE   & 0.3950 & 0.4522 & 0.0443 & 0.0049 & 0.0151 & 0.0066 & 0.0780 & 0.0136 \\
		& LE    & 0.1104 & 0.4570 & 0.0113 & 0.0008 & 0.0082 & 0.0024 & 0.0065 & 0 \\
		& UMAP   & \textbf{0.6819} & 0.7376& 0.0044 & \textbf{0.0604} & 0.3145 & 0.0447 & 0.3065 & 0.0020 \\
		& Vec2vec & 0.5549 & \textbf{0.8093} & \textbf{0.0605} & 0.0200 & \textbf{0.5191} & \textbf{0.1085} & \textbf{0.3066} & \textbf{0.1080} \\
		\bottomrule
	\end{tabular}%
	\label{tab:imageClutering}%
\end{table}%

To test the performance of \emph{Vec2vec}, we compare the performances in the eight datasets as shown in Table~\ref{tab:imageClutering}. We can find that \emph{Vec2vec} gets the best performance in the ``COIL20" and ``CIFAR-10" datasets, while UMAP gets the best performance in the other two image datasets. For the ``CIFAR-10" and ``SVHN" dataset, the ARI results of all seven methods are very small. We can find that \emph{Vec2vec} gets the best ARI results in all the four text datasets. 

We assume that the results are in Gaussian distribution and employ the Student's t-test to test the significant difference between the compared methods in the eight datasets. For UMAP, the H-value and p-value are 0 and 0.3118. \textbf{Therefore, UMAP is as good as \emph{Vec2vec} in statistics and the performances of the two methods have no significant difference in the eight datasets.} For the local LLE and LE methods, the H-values are both 1 and the p-values are 0.0188 and 0.0097. \textbf{Therefore, \emph{Vec2vec} is significantly better than LLE and LE in the eight datasets in statistics.} For the global PCA, CMDS, and Isomap method, the p-values are 0.0243, 0.0141, and 0.0411. The H-values are all 1. \textbf{Therefore, \emph{Vec2vec} is significant better than the global PCA, CMDS, and Isomap in the eight datasets in statistics.} The experimental results show that skip-gram model is very effective to obtain the embedding while preserving the similarity of data points, which are important for clustering.

\subsection{Parameter Sensitivity}~\label{parameter}

The \emph{Vec2vec} algorithm involves several parameters. We examine how the different choices of the target dimensionality $d$ and the number of $topk$ neighbors affect the performance of \emph{Vec2vec} on the four typical image datasets. We perform 4-fold cross-validation and employ KNN as the classifier. We utilize the accuracy score as the evaluation metric. In this experiments, except for the parameter tested, all other parameters are set to be their default values.

\begin{figure}[!t]
	\centering
	\subfigure[Accuracies with the change of target dimensionality]{%
		\label{1a}
		\includegraphics[width=5.5cm]{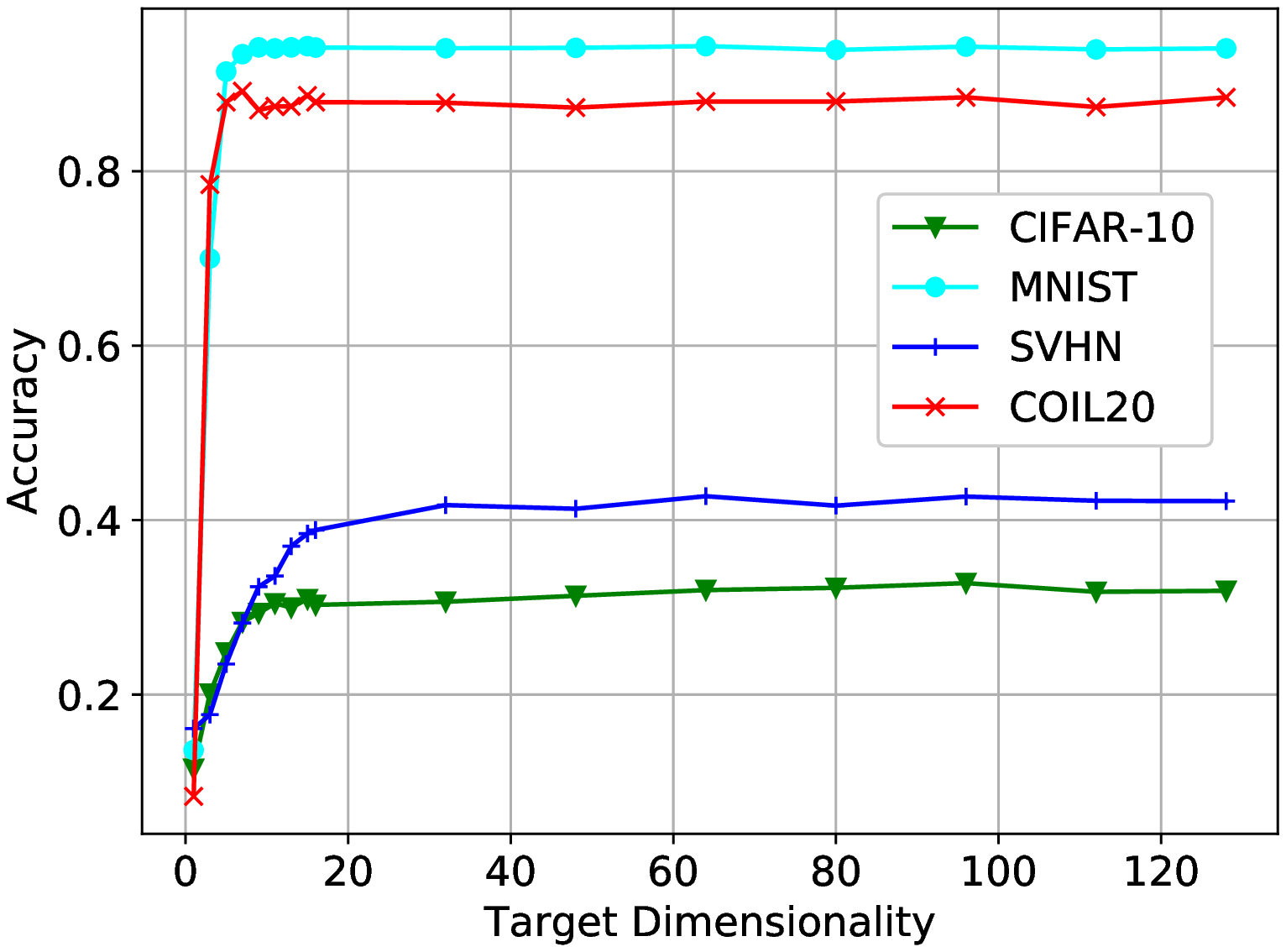}
	}
	\quad
	\subfigure[Accuracies with the change of the number of neighbours]{%
		\includegraphics[width=5.5cm]{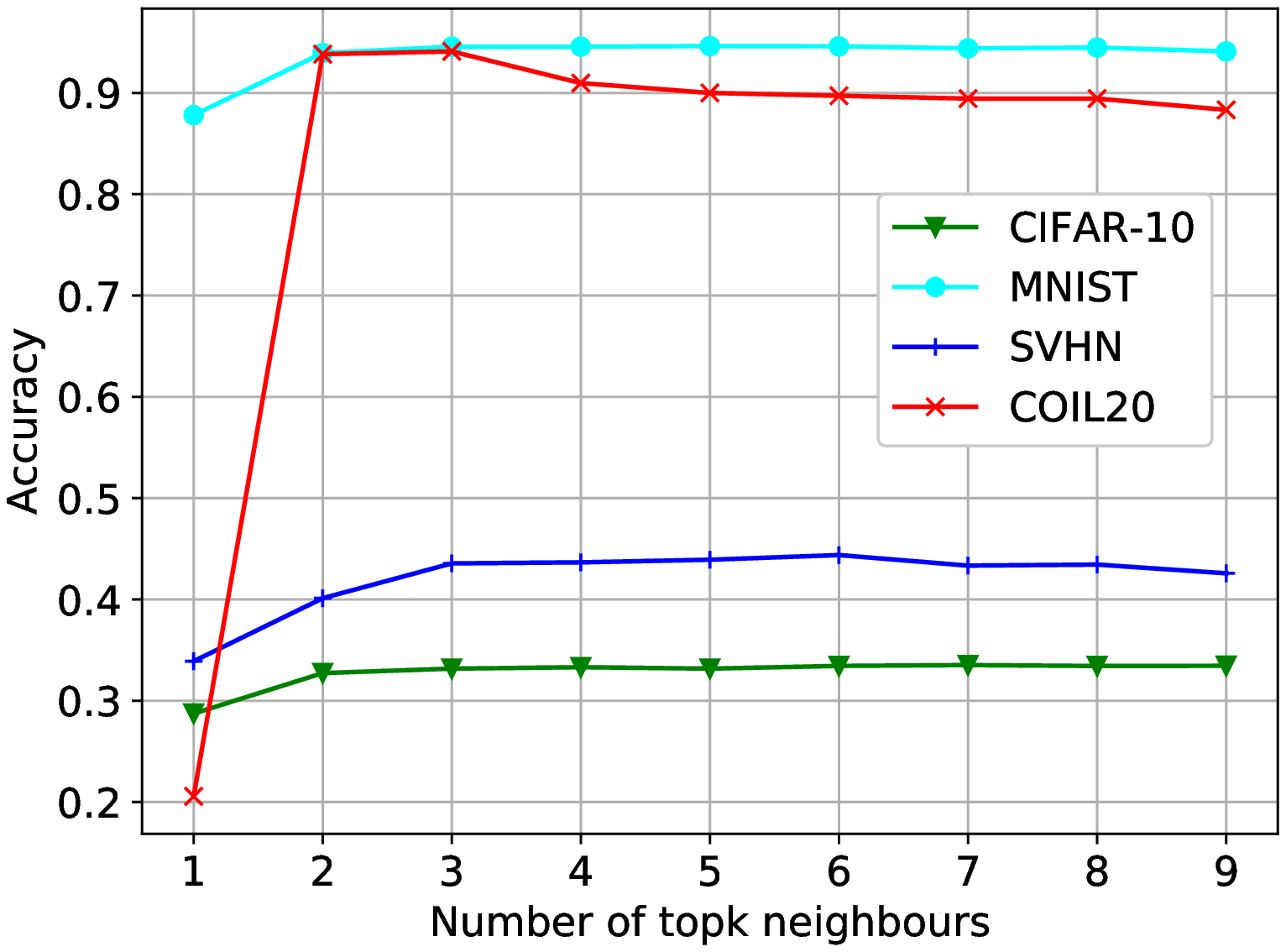}
		\label{1b}
	}
	\caption{Testing the parameter sensitivity of \textit{Vec2vec} with data classification.}
	\label{figParameters}
\end{figure}

Figure~\ref{figParameters} shows the performance evaluation of \emph{Vec2vec} with the change of the two parameters. For the target dimensionality $d$, \textit{Vec2vec} gets its best performance when $d$ is approximately equal to the number of classes in a dataset, which is the true dimensionality of the dataset. For the parameter of the number of $topk$ neighbors in building the similarity graph, \textit{Vec2vec} gets its best performance when $topk$ is less than 5 in all the four datasets. It shows that the neighborhood similarity graph is sparse and \textit{Vec2vec} is computationally efficient. \textbf{We can find that the performance of \emph{Vec2vec} is stable when the two parameters reach a certain value. It is important to find that since it is easy to choose the parameters of \emph{Vec2vec} in real applications.}

\section{Conclusion}~\label{conclusion}

In this paper, we study the local nonlinear dimensionality reduction to relieve the curse of dimensionality problem. To reduce the computational complexity, we generalize the skip-gram model for representation learning of words to matrices. To preserve the similarities between data points in a matrix after dimensionality reduction, we select the neighbors of the data points to establish a neighborhood similarity graph. We raise a hypothesis that similar data points tend to be in similar contexts in the feature space, and define the contexts as the co-occurrences of data points in the sequences of random walks in the neighborhood graph.

We analyze the computational complexity of \emph{Vec2vec} with the state-of-the-art local dimensionality reduction method UMAP. We find that our \emph{Vec2vec} is efficient in datasets with both a large number of data samples and high-dimensionality, while UMAP can be scalable to datasets with a large number of data samples, but it is sensitive to high dimensionality of the data. We do extensive experiments of data classification and clustering on eight typical real-world datasets for dimensionality reduction to evaluate our method. Experimental results show that \emph{Vec2vec} is better than several classical dimensionality reduction methods and is competitive with recently developed state-of-the-arts UMAP in the statistical hypothesis test. 

\subsubsection*{Acknowledgment}
We are thankful to the anonymous reviewers. This research is partially supported by the National Key R\&D Program of China (Grant No. 2018YFB0203801), National Natural Science Foundation of China (Grant No. 61802424, 61702250), and ICT CAS research grant CARCHB202017.

\bibliographystyle{splncs04}
\bibliography{dasfaa2021}

\end{document}